\documentclass[11pt]{article}
\usepackage{xspace}
\newcommand{\model}{Traffic-R1\xspace}

\usepackage[final]{acl}

\usepackage{times}
\usepackage{latexsym}

\usepackage[T1]{fontenc}
\usepackage[utf8]{inputenc}

\usepackage{microtype}
\usepackage{enumitem}
\usepackage{booktabs}
\usepackage{multirow}
\usepackage{tabularray}
\usepackage{makecell}
\usepackage{inconsolata}
\usepackage{amsmath}
\usepackage{graphicx}

\usepackage{multirow}
\usepackage{graphicx}
\usepackage[normalem]{ulem}
\usepackage{tcolorbox}
\tcbuselibrary{theorems}
\usepackage{amsmath}
\usepackage{amsthm}
\usepackage{microtype}
\usepackage{amssymb}
\usepackage{times}
\usepackage{mathtools}
\usepackage{algorithm}
\usepackage{algpseudocode}
\useunder{\uline}{\ul}{}

\usepackage{color}
\usepackage{hyperref}

\definecolor{tblue}{RGB}{31,119,180}
\definecolor{torange}{RGB}{255,127,14}
\definecolor{tgreen}{RGB}{44,160,44}
\definecolor{tred}{RGB}{214,39,40}
\definecolor{tpurple}{RGB}{148,103,189}
\definecolor{lightgreen}{RGB}{0, 135, 125}
\definecolor{lightred}{RGB}{255, 87, 51}

\def\model{Traffic-R1\xspace}
\newcommand{\paratitle}[1]{\noindent\textbf{#1}}

\newtheorem{myDef}{Definition}[section]

\definecolor{mycolor}{RGB}{70,130,180}
\newtcbtheorem[auto counter, number within = section]{cmt}{}{
 colbacktitle = mycolor, colframe = mycolor,
 colback = mycolor!10!white,
 fonttitle=\bfseries\small,
 fontupper=\small, % 设置大写字母的字体为小字体
  fontlower=\small, % 设置小写字母的字体为脚注字体
 left=1mm, % 左边距
  right=1mm,
}{t}

\title{Traffic-R1: Reinforced LLMs Bring Human-Like Reasoning to Traffic Signal Control Systems}

\author{
    \textbf{Xingchen Zou\textsuperscript{\rm 1,2}},
    \textbf{Yuhao Yang\textsuperscript{\rm 3}},
    \textbf{Zheng Chen\textsuperscript{\rm 2}},
    \textbf{Xixuan Hao\textsuperscript{\rm 1}}, 
    \\ 
    \textbf{Yiqi Chen\textsuperscript{\rm 2}}, 
    \textbf{Chao Huang\textsuperscript{\rm 3}},
    \textbf{Yuxuan Liang\textsuperscript{\rm 1}}\thanks{Corresponding author. Email: yuxliang@outlook.com.}
    \\
$^1$The Hong Kong University of Science and Technology (Guangzhou), \\ $^2$PCITECH,
  $^3$The University of Hong Kong\\
\texttt{\{xzou428,xhao390\}@connect.hkust-gz.edu.cn, \{chaohuang75\}@gmail.com, }\\
\texttt{\{chenzheng1, chenyiqi\}@pcitech.com, \{yuhao-yang, yuxliang\}@outlook.com}
}
\begin{document}
\maketitle
\begin{abstract}
\vspace{-1em}
We introduce \model, a 3B-parameter foundation model with human-like reasoning for Traffic signal control (TSC), developed via self-exploration and iterative reinforcement of LLM with expert guidance in a simulated traffic environment.
Compared with traditional reinforcement learning and recent LLM-based methods, \model offers three main advantages: zero-shot generalization, transferring unchanged to new road networks and out-of-distribution incidents by leveraging internal traffic-control policies and reasoning; a compact 3B-parameter design that supports real-time inference on mobile-class chips for edge deployment; and an explainable TSC process that enables multi-intersection coordination through communication and an asynchronous communication network.
Extensive benchmarks show \model outperforms strong baselines and training-intensive RL controllers. In production, the model now manages signals affecting over 55,000 drivers daily, reduces average queue lengths by more than 5\%, and halves operator workload.  Our model
is available at \href{https://huggingface.co/Season998/Traffic-R1}{https://huggingface.co/Season998/Traffic-R1}.
\vspace{-1em}
\end{abstract}

\vspace{-0.5em}
\section{Introduction}
\vspace{-0.5em}
\label{sec:intro}

Rapid urbanization and surging vehicle ownership intensify congestion, wasting billions of productive hours, burning vast fuel reserves, and driving nearly a quarter of urban greenhouse emissions. Prolonged delays raise crash rates, slow emergency response, exacerbate cardiopulmonary pollution, and unfairly burden transit-poor communities. Within this broad societal context, traffic-signal control (TSC), which coordinates phase sequences and durations at signalized intersections, remains a principal lever for mitigating congestion and improving network throughput \cite{yau2017survey,ye2019survey,wei2021recent,zhang2024trafficgpt}.

\begin{figure}[t]
    \centering
    \includegraphics[width=1.05\linewidth]{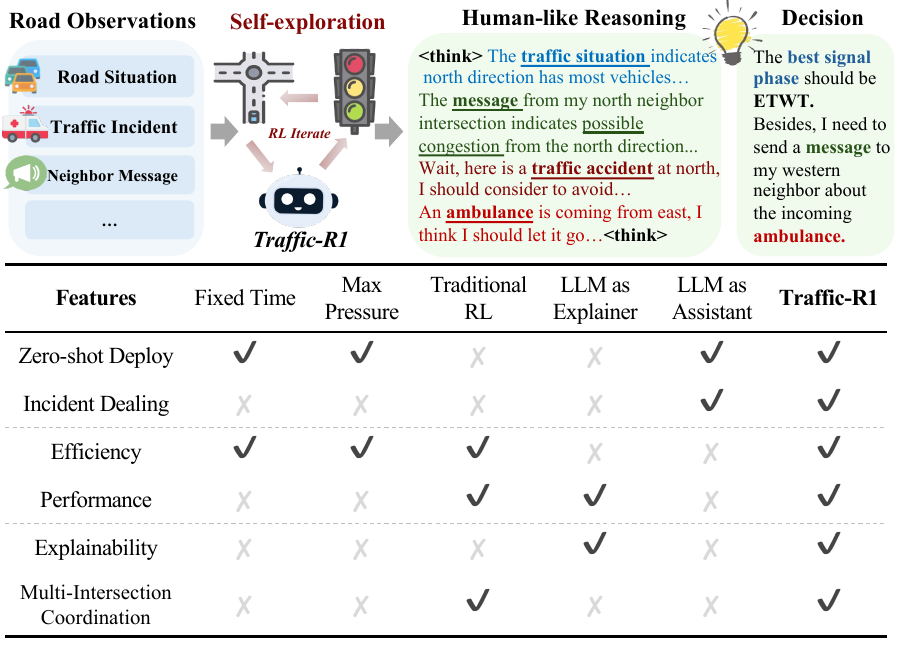}
    \vspace{-2em}
    \caption{Introduction of Traffic-R1, a foundation (covering six features) reinforced LLM for TSC systems.}
    \label{fig:intro}
    \vspace{-1.5em}
\end{figure}

Traditional controllers like FixedTime \cite{koonce2008traffic} and MaxPressure \cite{varaiya2013max} rely on fixed heuristics and thus adapt poorly to fluctuating demand. Reinforcement learning (RL) replaces these hand-crafted rules with a data-driven policy: each cycle observes lane queues, delays, and neighboring signal states, selects a phase (or duration) as the action, and receives a reward linked to delay reduction or throughput gain. Deep RL further augments this paradigm by learning the policy end-to-end with expressive function approximators. These advances achieve impressive benchmark scores in simulation \cite{srinivasan2006neural,zhao2011computational}. However, field deployment remains rare as existing methods still suffer from
(i) cross-region generalization: policies tuned on one city transfer poorly to another \cite{zhao2024tsclip,wei2021recent};
(ii) interpretability: nontransparent decisions undermine practitioner trust \cite{vouros2022explainable,glanois2024survey}; and
(iii) robustness to out-of-distribution (OOD) events: models falter during incidents or emergency-vehicle priority scenarios \cite{glanois2024survey,wei2018intellilight}.

Recently, large language models (LLMs) have been enlisted to alleviate these shortcomings. Two integration paradigms dominate. In the \textbf{LLM Explainer} paradigm \cite{lai2023llmlight,yuan2025collmlight}, an LLM is trained to verbalize the policy of an RL controller, translating opaque action choices into natural language rationales. In contrast, the \textbf{LLM assistant} framework \cite{wang2025vlmlight,wang2024llm} keeps the RL agent in charge of routine control and consults an LLM only when OOD incidents arise. \emph{Both paradigms are promising yet remain distant from large-scale deployment:} LLM Explainers inherit the coverage and performance ceiling of the underlying RL policy, and their post-hoc narratives can diverge from the controller's true internal logic \cite{wu2025llm,glanois2024survey,wei2018intellilight,agarwal2024faithfulness,malin2025review}; LLM Assistants introduce additional prompt engineering and repeated LLM queries, inflating latency and computation while providing limited benefit for everyday signal timing. Hence practitioners remain cautious about adopting current LLM-enhanced TSC solutions.

To date, operational TSC systems still rely on heuristic rule sets and substantial human oversight to cope with routine flow and unexpected incidents \cite{mandhare2018intelligent,muller2021towards,mahavar2018literature}. Bridging this research-deployment gap requires a \emph{foundational model} for TSC systems, i.e., a single, versatile agent capable of (1) \textbf{zero-shot generalization} to unseen traffic networks and OOD incidents, (2) \textbf{resource-efficient inference} on edge hardware such as mobile platforms, and (3) \textbf{human-like \& transparent reasoning} that supports explainable decision-making and multi-intersection coordination. Figure \ref{fig:intro}) schematically illustrates these three desiderata in the envisioned foundational agent.

We answer this call with \textbf{Traffic-R1},  a lightweight reinforced LLM with human-like reasoning, designed as a foundational traffic controllers incorporating the six key features from Figure \ref{fig:intro}. Built on the efficient Qwen2.5-3B, Traffic-R1 employs a two-stage agentic RL finetuning approach: an offline stage integrates human expert knowledge from TSC recordings, and an online stage adapts the model to dynamic scenarios through exploration in simulated environments.
Inspired by recent studies~\cite{bai2022training,liu2024deepseek,yu2025dapo,zhou2310vision}, our RL framework trains Traffic-R1 to generate Chain-of-Thought (CoT) reasoning along with actions via self-iteration. The learning process is guided by a policy-based reward model that incorporates rewards for both output format and action correctness. In the offline stage, action rewards are calculated based on alignment with human expert decisions; in the online stage, they are derived from simulated traffic feedback. This training scheme enables the lightweight model to develop robust reasoning through self-exploration, leading to strong zero-shot TSC performance and generalization to out-of-distribution (OOD) scenarios, all while maintaining resource-efficient inference.

By generating its own training samples within our RL framework, Traffic-R1 mitigates the catastrophic forgetting often observed when smaller models are finetuned on synthetic data from larger ones. As a result, it retains strong general language communication skills and explainable reasoning. o leverage these language skills for multi-agent traffic coordination, we introduce an asynchronous communication network that enables LLM agents to coordinate across multiple intersections via message-passing, mimicking the collaboration of human traffic agents.

In summary, our contributions are:
\begin{itemize}[leftmargin=*,topsep=0em]
    \item \textbf{Foundation model for TSC systems}: We present the first LLM-based, general-purpose controller that can operate at any interaction without additional training, handling routine signaling, incident management, and emergency-vehicle prioritization with human-level reasoning.
\vspace{-2em}
    \item \textbf{Lightweight yet high-performing}: Our two-stage RL training yields a 3B-parameter LLM that matches or exceeds much larger models (e.g., GPT-4o) and strong expert baselines, while remaining suitable for inference on edge devices.
\vspace{-2em}
    \item \textbf{Human-like reasoning and communication}: We achieve human-like reasoning for explainable TSC through self-iteration of the reinforced LLM. This reasoning, combined with language capabilities, is further utilized in our asynchronous communication network for effective coordination across multiple intersections.
\vspace{-1em}
    \item \textbf{Extensive validation and field deployment}: Evaluated on standard TSC benchmarks and out-of-distribution incident tasks, achieving stable state-of-the-art performance. In live deployment serving over 55,000 drivers daily, trials show a >5\% reduction in average queues and a >50\% reduction in operator workload for phase planning and incident response.
\end{itemize}

\vspace{-0.5em}
\section{Related Works}
\vspace{-0.5em}
\label{sec:relate}
\textbf{Traffic Signal Control.} 
TSC is essential for traffic management. The FixedTime method, one of the earliest and most widely used approaches, relies on predetermined cycle lengths and phase allocations set by human experts for each intersection~\cite{yau2017survey,serafini1989mathematical}. It is simple and stable but inefficient under dynamic traffic and requires substantial human effort~\cite{serafini1989mathematical,thunig2019optimization}. Maxpressure introduces adaptive control by prioritizing movements based on pressure, defined from queue lengths per direction~\cite{varaiya2013max,mercader2020max}, yet it and FixedTime remain constrained by fixed rules that struggle with varying scenarios~\cite{rasheed2020deep,wei2021recent,abdulhai2003reinforcement}.
Machine learning has produced RL-based TSC methods such as CoLight~\cite{wei2019colight}, CosLight~\cite{ruan2024coslight}, and MPLight~\cite{chen2020toward}, which improve performance in simulation but often fall short for real-world deployment~\cite{chen2022real,wei2021recent,qadri2020state}. Recent work explores integrating large language models into TSC, but significant challenges remain before these approaches are practical for field use~\cite{lai2023llmlight,yuan2025collmlight,wang2025vlmlight,wang2024llm,wen2023dilu,zhao-etal-2025-urbanvideo,feng2025citybench}.

\noindent
% \textbf{LLM for Traffic Signal Control.} 
% The development of LLMs has shown the potential for foundation models across various domains~\cite{zou2025deep,yang2024graphagent,huang2025towards,liu2025citylens,wang2024graph,lai2025ustbench,hao2025unlocking} due to their strong generalization capabilities and human-like reasoning. Recent research has investigated integrating LLMs to enhance TSC systems~\cite{lai2023llmlight,yuan2025collmlight,wang2025vlmlight,wang2024llm}. These efforts can be categorized into two paradigms:
% (1) LLM as Explainer: Represented by LLMLight~\cite{lai2023llmlight}, this approach uses GPT-4 to generate explanations for actions selected by reinforcement learning (RL)-based TSC methods and fine-tunes a smaller LLM to replicate these synthetic textual explanations.
% (2) LLM as Assistant: Represented by LA-Light~\cite{wang2024llm}, this paradigm employs an LLM as an additional assistant to handle traffic incident after applying conventional TSC methods, such as Maxpressure.
% While these paradigms highlight the potential of LLMs in TSC, the industry remains cautious about their adoption in real-world TSC systems due to limited working scenarios and cost-benefit considerations. Neither the explainer nor the assistant paradigm provides benefits that justify the additional deployment and computational costs, as LLMs in these methods only working as imitators or supplements to traditional TSC methods. For real-world TSC deployment, a new paradigm is needed, where LLMs act as human-like agents to deliver comprehensive, general and labor-saving control.

\noindent
\textbf{RL finetuning for LLM.} 
LLMs often require fine-tuning to adapt to specific tasks or human preferences~\cite{wu2025llm,minaee2024large,ding2025understanding,li2025urban}. Instruction finetuning~\cite{chung2024scaling,dettmers2023qlora} is widely used because it is simple and effective for task- and format-specific adaptation. RL finetuning, particularly RL from Human Feedback~\cite{ouyang2022training}, has seen less adoption for downstream use due to its complex pipeline and high compute cost\cite{gao2023scaling,bai2022training,ramamurthy2022reinforcement}. Recent work shows LLMs can develop reasoning through interaction with RL environments, and \citet{liu2024deepseek} reduces compute needs by using group policy-based optimization. RL finetuning offers practical advantages: (1) self-exploration reduces dependence on large labeled datasets, (2) it guides models to acquire capabilities rather than imitate or memorize outputs, and (3) KL-divergence constraints on updates help mitigate catastrophic forgetting by anchoring the policy to the pretrained distribution\cite{cao2024survey,xu2025towards,ke2025survey}.

\vspace{-0.5em}
\section{Preliminaries}
\vspace{-0.5em}
\label{sec:model}

\begin{myDef}
\textbf{Road Network.} The road network is modeled as a directed graph with intersections $\mathcal{V}$ and lanes $\mathcal{L}$. Lanes are classified into three types: (1) go-through lanes ($\mathcal{L}_{\text{go}}$), (2) left-turn lanes ($\mathcal{L}_{\text{left}}$), and (3) right-turn lanes ($\mathcal{L}_{\text{right}}$). Each lane connects to neighboring intersections and is divided into segments $S = \{s_1, \ldots, s_n\}$ based on their distance from the intersection.
\end{myDef}
\vspace{-0.5em}
\begin{myDef}
\textbf{Signal Phase.} At each signal-switching time step, the model assigned to an intersection selects a signal phase from a predefined set $\mathcal{A} = \{a_1, \ldots, a_m\}$. A signal phase is defined as $a = \text{set}(\mathcal{L}_{\text{allow}})$, where $\mathcal{L}_{\text{allow}}$ represents the set of lanes permitted to proceed without conflicting movements (i.e., green light for $\mathcal{L}_{\text{allow}}$ and red light for conflicting lanes).
\end{myDef}
\vspace{-0.5em}
\begin{myDef}
\textbf{Traffic Signal Control System.} The traffic signal control system comprises multiple agents $\Pi = \{\pi_1, \ldots, \pi_n\}$, each managing signal control at one of $n$ intersections in a road network. Each agent $\pi_i$ collaborates with neighboring agents through traffic observations and message passing at signal-switching time steps to coordinate multi-intersection operations, such as green wave synchronization and emergency response.
\end{myDef}

\section{Methodology}
\label{sec:solution}
\vspace{-0.6em}
\begin{figure}[!b]
    \centering
    \vspace{-1em}
    \includegraphics[width=1.0\linewidth]{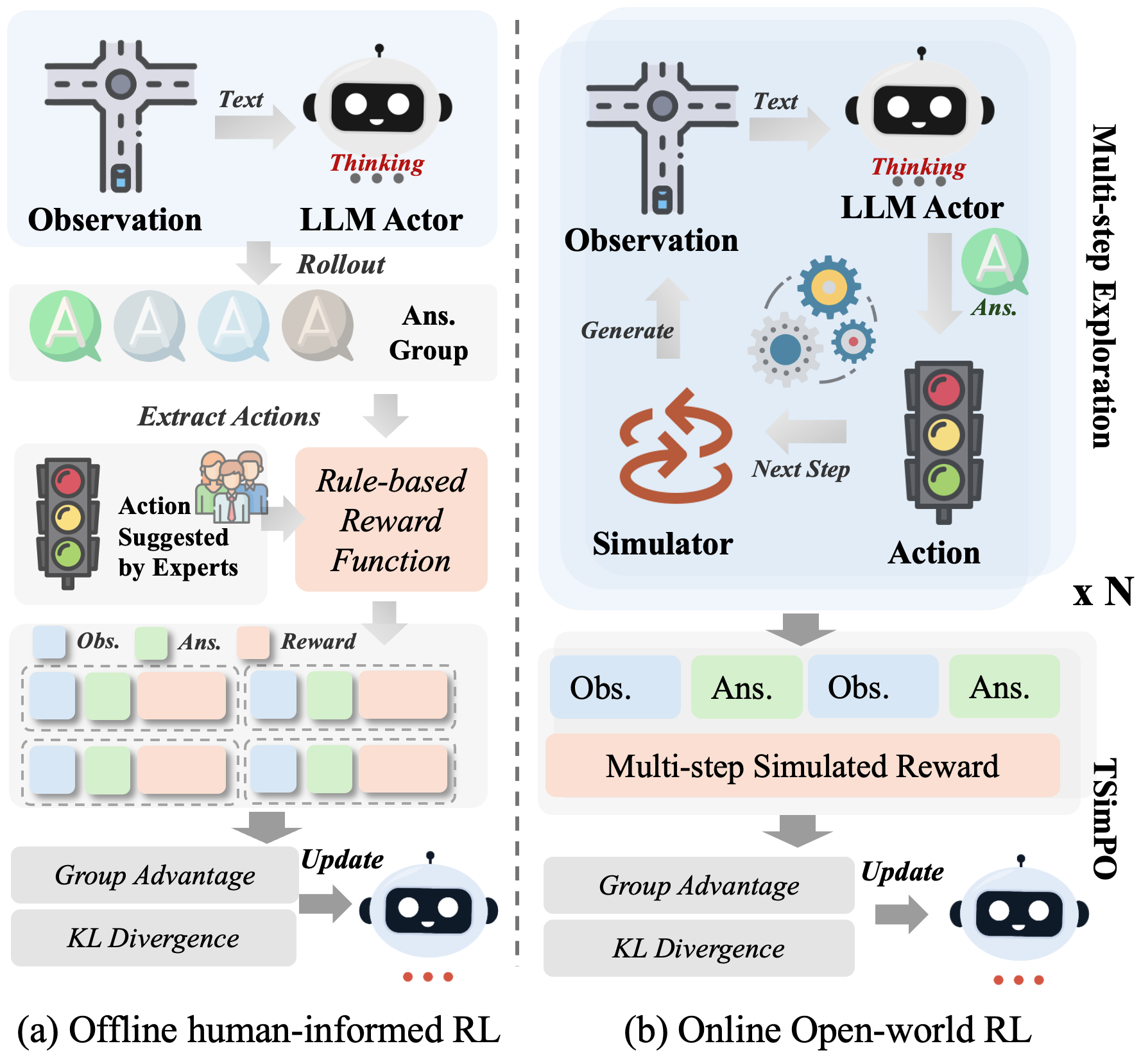}
    \vspace{-2em}
    \caption{Introduction of the two-stage RL framework}
    \label{fig:frame}
\end{figure}

In this section, we describe the training pipeline for \model. As shown in Figure \ref{fig:frame}, we utilize a two-stage RL framework comprising offline human-informed RL and online open-world RL to finetune the LLM. For each stage, we define distinct training datasets, environments, policy optimization methods, and reward designs to support their functionality. Besides, we propose an asynchronous communication network for reinforced LLM to support efficient multi-intersection coordination.

\vspace{-0.5em}
\subsection{Human-Informed RL}
Existing LLMs for TSC are typically finetuned to imitate traditional RL models. This approach has two major flaws: first, finetuning LLMs to imitate traditional RL models confines their performance and generalization to the capabilities of the teacher RL model. Second, action trajectories produced by RL models through iterative optimization in simulated environments may not provide clear reasoning and logic for LLMs, as some RL decision policies and actions can be suboptimal or impractical for general TSC, focusing instead on narrow performance metrics. 
To address these limitations, we propose a human-informed RL finetuning stage that incorporates an expert-collaborative TSC dataset and offline policy optimization. This approach replaces the RL model teacher with real human traffic experts to guide the finetuning of the LLM.
\vspace{-0.5em}
\subsubsection{Expert-Collaborative TSC Dataset}

Our RL finetuning approach reduces the training data requirement from hundreds of thousands of samples typically needed for instruction finetuning to just thousands. This enables the creation of a TSC dataset with actions provided by human experts for each traffic scenario. We developed a 3,000-sample dataset using a human-in-the-loop pipeline. For each traffic scenario, a base model suggests an action. This action is then validated for effectiveness using the SUMO simulator and reviewed by two human traffic control experts. If an action is rejected, experts provide the correct one. This process, involving 11 experts, resulted in our final dataset.
Crucially, we only include the final, validated actions, excluding any reasoning steps. This design encourages the model to develop its own reasoning capabilities during RL finetuning, rather than imitating an external thought process.
\begin{figure}[t]
\centering
\vspace{-1em}
\includegraphics[width=1.0\linewidth]{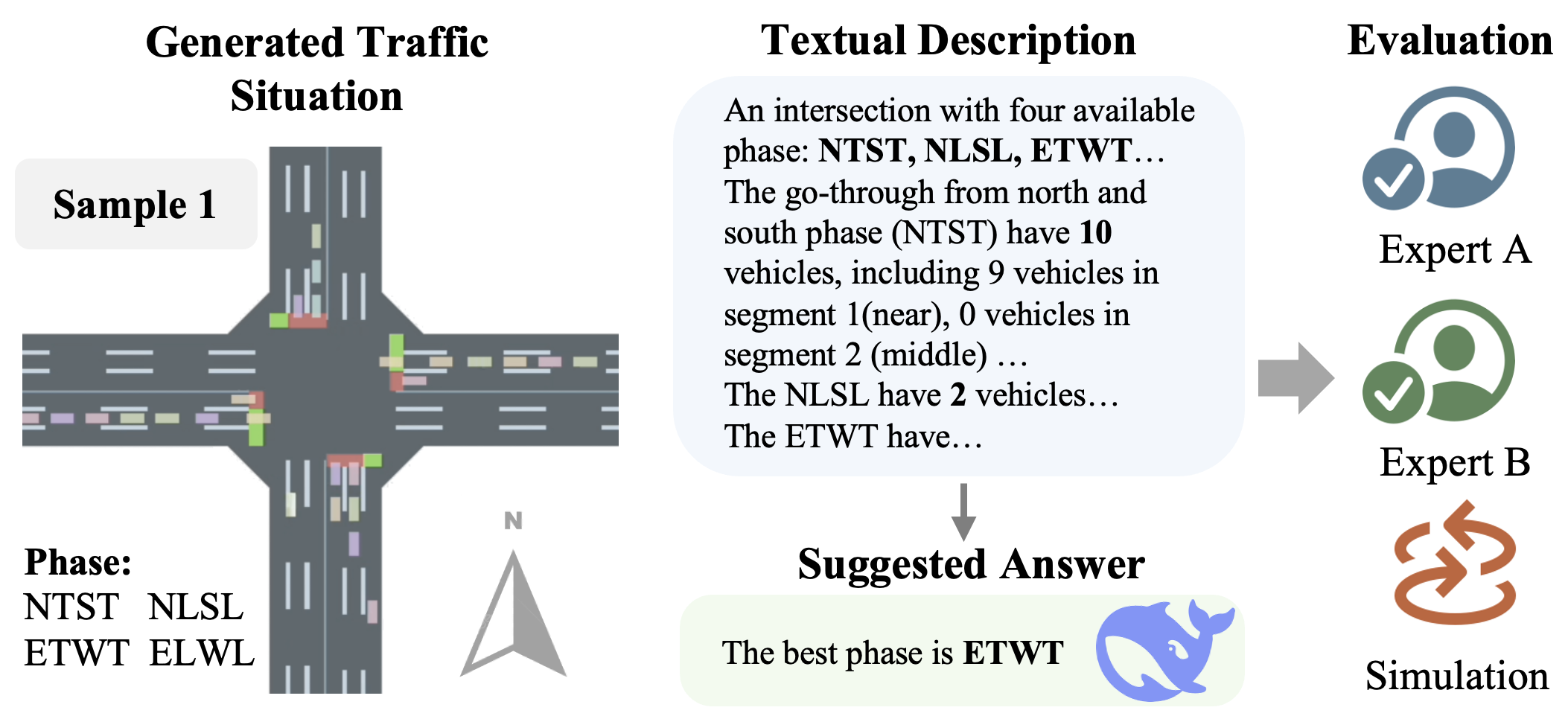}
\vspace{-2em}
\caption{Expert-Collaborative Dataset Construction}
\vspace{-1.5em}
\label{fig:dataset}
\end{figure}

\vspace{-0.5em}
\subsubsection{Offline Policy Optimization}

Inspired by the work of \cite{liu2024deepseek}, which effectively finetunes LLMs on mathematics and coding datasets containing only final answers or actions without reasoning text, our approach uses the same offline RL framework to finetune the LLM on our expert-collaborative TSC dataset to promote self-thinking in traffic control. The core process involves an LLM policy model interacting with an offline environment, guided by a rule-based reward derived from expert-provided answers.  Given a textual prompt $x$ describing a specific traffic scenario, the LLM, parameterized by $\theta$, generates an output sequence $y$ autoregressively according to its policy $\pi_\theta(y \mid x) = \prod_{t=1}^{|y|} \pi_\theta(y_t \mid x, y_{<t})$.

\paratitle{Input Template for Rollout.} Rollout is a critical component of RL iterations, which involves using the original LLM $\pi_\theta$ to produce a variety of structured responses. To guide the interaction process, the LLM is prompted with specific templates that include format instructions, ensuring the generated sequence $y$ contains both task-specific reasoning and answer components in a structured, extractable format for the offline reward policy. The input prompt template is presented in Appendix \ref{app:prompt}.

\paratitle{Offline Rule-based Reward.} By treating the offline TSC task as a math-like problem-solving process during our offline RL stage, the rule-based reward is required to be clear and simple, minimizing computational complexity and preventing reward hacking. The reward function $R$ is a weighted combination of an accuracy reward $R_{\text{acc}}$ and a format reward $R_{\text{format}}$, defined as: $R(x,y) \in [0,1] = \begin{bmatrix} w_{\text{acc}}, w_{\text{format}} \end{bmatrix} \cdot \begin{bmatrix} R_{\text{acc}}, R_{\text{format}}\end{bmatrix}^T$. Since signal actions are mutually exclusive, $R_{\text{acc}} = 1$ only when the generated action exactly matches the expert-suggested action. Similarly, $R_{\text{format}} = 1$ only when the generated sequence $y$ fully adheres to the specified reasoning and answer format instructions.

\paratitle{Reinforcing Reasoning via Policy Optimization.} To optimize the policy $\pi_\theta$ with parameters $\theta$ in an offline environment, we employ Group Relative Policy Optimization (GRPO)~\cite{liu2024deepseek} to ensure stable gradient updates.
Let $\pi_\theta$ represent the policy and $\{a_i^{(j)}\}_{j=1}^k$ denote each input prompt $x_i$ is paired with $k$ candidate completions sampled from the current policy. The reward function $R$ assigns a score $r_i^{(j)}$ to each completion $a_i^{(j)}$.
To address high variance in policy gradients, GRPO computes group-normalized advantages for each completion $a_i^{(j)}$ generated from the same input $x_i$, as shown in Equation \ref{equ:adv}. This approach centers the rewards within each group, mitigating the impact of absolute reward magnitudes:
\begin{equation}
\setlength\abovedisplayskip{-1pt}%shrink space
\setlength\belowdisplayskip{-1pt}
\label{equ:adv}
A_i^{(j)} = r_i^{(j)} - \frac{1}{k} \sum_{l=1}^k r_i^{(l)},
\end{equation}
\vspace{-1em}
The policy is updated by maximizing the clipped surrogate objective:
\begin{equation}
{\small
\setlength\abovedisplayskip{-2pt}%shrink space
\setlength\belowdisplayskip{0pt}
\label{equ:grpo}
\begin{split}
\mathcal{L}(\theta) = \mathbb{E} \Bigg[ & \min \bigg( \rho_i^{(j)} A_i^{(j)}, \text{clip}(\rho_i^{(j)}, 1 - \epsilon, 1 + \epsilon)* \\
& A_i^{(j)} \bigg) - \beta \mathbb{D}_{\text{KL}} \left[ \pi_\theta(\cdot|x) \parallel \pi_{\theta_{\text{ref}}}(\cdot|x) \right] \Bigg],
\end{split}
}
\end{equation}
\vspace{-1em}

where $\rho_i^{(j)} = \frac{\pi_\theta(a_i^{(j)} | x_i)}{\pi_{\theta_{\text{ref}}}(a_i^{(j)} | x_i)}$ is the likelihood ratio between the current policy $\pi_\theta$ and the reference policy $\pi_{\theta_{\text{ref}}}$, and $\epsilon$ is the clipping threshold. The expectation $\mathbb{E}[\cdot]$ is computed over $(x_i, a_i^{(j)})$ drawn from $\pi_{\theta_{\text{ref}}}$.The coefficient $\beta$ determines the strength of the Kullback-Leibler divergence penalty $\mathbb{D}_{\text{KL}} \left[ \pi_\theta \parallel \pi_{\theta_{\text{ref}}} \right]$. In practice, reference policy $\pi_{\theta_{\text{ref}}}$ is typically set to a snapshot of the previous policy, which stabilizes training to inspire deep thinking instead of imitation by constraining policy updates.

\subsection{Open-World Reinforcement Learning}
Although the human-informed offline RL has finetuned LLM to learn from human experts for stable performance on TSC tasks, the model's capacity is limited to the expert knowledge extracted from the dataset. In this section, we propose an open-world online RL to inspire LLM explore multi-step and multi-intersection TSC networks. This approach allows LLM to interact with the online dynamic simulated traffic environment and update its policy based on online reward for better performance.

\paratitle{Online Traffic Simulation} 
To simulate the multi-intersection and multi-step dynamics of real-world traffic flow, we constructed a $4 \times 4$ simulated road network with 300-meter roads between each intersection. The 16 positions in the network represent most typical road scenarios encountered at real-world intersections. Traffic flow within the network is randomly generated, allowing for up to 8,000 vehicles over the course of one hour. For efficiency in iterations, we utilize CityFlow~\cite{zhang2019cityflow} as the simulator to model the traffic dynamics resulting from the actions of the LLM. The online multi-step rewards $R_{traj}$ are quantified based on the cumulative average queue length and waiting time caused by a series of multi-intersection actions. We use group advantage as final rewards to mitigate random variations in environment during online training through group mean comparison, ensuring stable optimization gradients.

\paratitle{Multi-step Policy Optimization}
Existing RL finetuning typically uses offline, single-turn settings (e.g., math). Traffic online RL requires LLMs to act in interactive, multi-step environments with stochastic feedback. Methods~\cite{ragen} that concatenate observations and model responses into a single trajectory with a trajectory-level reward are impractical for TSC: context lengths become very large, raising compute and memory costs and diluting token-level attention on critical observations and actions. Besides, traffic is also partially observable and not well modeled as a perfect Markov decision process, so directly porting trajectory-concatenation approaches from digital games will likely produce brittle policies that generalize poorly and offer limited interpretability, similar to traditional RL controllers.

To address the chaotic dynamics in real-world traffic systems, where the link between actions and subsequent states is not strictly continuous, we propose \textit{Stepwise Trajectory Policy Optimization} (STPO). This method assigns the total trajectory reward to each individual observation-action pair ($o_t, a_t$), decomposing the trajectory to provide denser, step-level reward signals and reduce computational overhead.
The step-wise reward $r_t$ is uniformly distributed from the total trajectory reward $R_{\text{traj}}$ over $T$ steps:
\begin{equation}
\setlength\abovedisplayskip{1pt}%shrink space
\setlength\belowdisplayskip{1pt}
    r_t = {R_{\text{traj}}(o_{1:T}, a_{1:T})}\cdot{T}^{-1},
\end{equation}
where $R_{\text{traj}}$ is the reward over the full sequence of observations $o_{1:T}$ and actions $a_{1:T}$.
The policy $\pi_\theta(a_t | o_t)$ is then optimized with the objective:
\begin{equation}
\label{equ:stpo_objective}
\setlength\abovedisplayskip{1pt}
\setlength\belowdisplayskip{1pt}
\mathcal{L}_{\text{STPO}}(\theta) = \mathbb{E}_{(o_t, a_t \sim \pi_{\theta_{\text{ref}}})} \left[ \log \pi_\theta(a_t | o_t) A_t \right],
\end{equation}
%\vspace{-1em}
The advantage calculation ($A_t$), Kullback-Leibler divergence penalty, and advantage clipping are analogous to those in the original GRPO formulation (see Equation~\ref{equ:grpo}).

\subsection{Asynchronous Communication Network}
Most TSC research assumes a synchronized parallel workflow, where all intersection agents act simultaneously. This simplification hinders real-world multi-intersection coordination, which is inherently asynchronous. Consequently, implementations in synchronized frameworks are often inefficient, requiring extensive shared global data, or ineffective, leading to incompatible decisions. While some reinforcement learning methods incorporate neighborhood observations, they typically suffer from poor generalization and scalability.

\begin{figure}[t]
    \centering
    \includegraphics[width=1.0\linewidth]{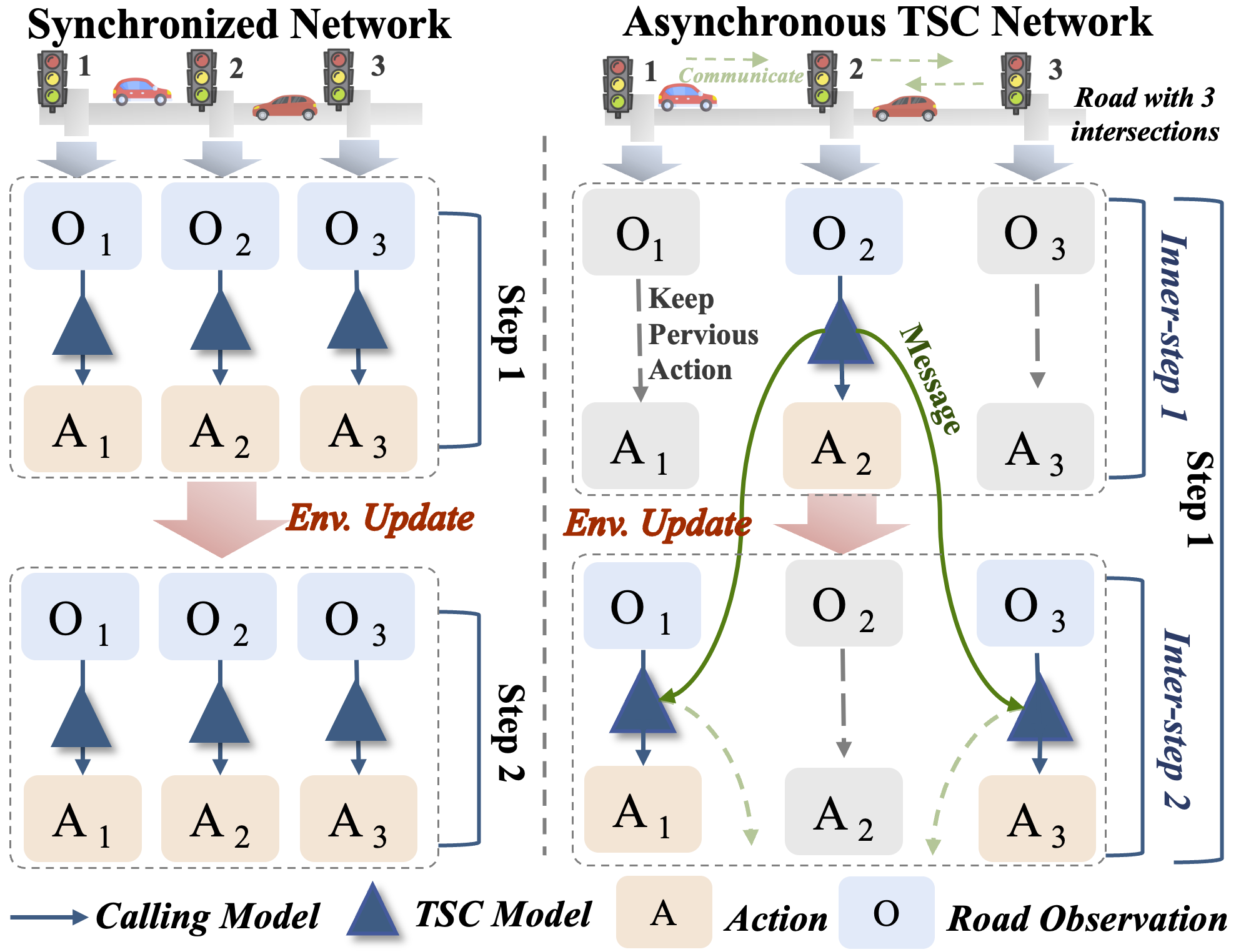}
    \vspace{-2em}
    \caption{Asynchronous communication network compared with conventional synchronized network}
    \label{fig:network}
    \vspace{-1.5em}
\end{figure}
To enable natural communication and unlock the potential of our LLM-based model, we propose an asynchronous communication network. As illustrated in Figure \ref{fig:network}, we partition intersections into two groups based on positional parity. Each TSC step is split into two inner steps: the first activates group one, and the second activates group two. This structure allows models activated in the first inner step to generate messages for their neighbors, which are then received and processed in the subsequent inner step. This framework fully utilizes the language capabilities of LLMs to enhance multi-intersection coordination. The pseudocode for this network is provided in Algorithm \ref{alg:async_symmetric_professional}.

\vspace{-0.5em}
\section{Experiments}
\vspace{-0.5em}

\begin{table}[b]
\centering
\small
\vspace{-1.5em}
\caption{Statistics of traffic flow datasets.}
\vspace{-1em}
\label{tab:dataset}
\resizebox{\columnwidth}{!}{%
\begin{tabular}{ccccccc}
\hline
\multirow{2}{*}{Dataset} & \multirow{2}{*}{Structure} & \multirow{2}{*}{Vehicles} & \multicolumn{4}{c}{Arrival rate (vehicles/5min)} \\ \cline{4-7} 
 &  &  & \multicolumn{1}{l}{Mean} & \multicolumn{1}{l}{Std} & \multicolumn{1}{l}{Max} & \multicolumn{1}{l}{Min} \\ \hline
Jinan1 & \multirow{3}{*}{$3 \times 4$} & 6295 & 523.67 & 98.52 & 671 & 255 \\
Jinan2 &  & 4365 & 362.83 & 74.81 & 493 & 236 \\
Jinan3 &  & 5494 & 456.92 & 160.87 & 569 & 362 \\ \hline
Hangzhou1 & \multirow{2}{*}{$4 \times 4$} & 2983 & 247.68 & 40.44 & 332 & 211 \\
Hangzhou2 &  & 6984 & 581.08 & 318.43 & 1145 & 202 \\ \hline
\end{tabular}%
}
\end{table}

% \usepackage{multirow}
% \usepackage{graphicx}
% \usepackage[normalem]{ulem}
\begin{table*}[]
\centering
\caption{Zero-shot performance comparison on conventional traffic signal control tasks (the smaller the better). The best results are in \textbf{bold} and second-best results are {\ul underlined}.}
\vspace{-1em}
\label{tab:exp}
\resizebox{\textwidth}{!}{%
\begin{tabular}{cccccccccccc}
\hline
\multirow{2}{*}{\textbf{Models}} & \multicolumn{2}{c}{\textbf{Jinan1}} & \multicolumn{2}{c}{\textbf{Jinan2}} & \multicolumn{2}{c}{\textbf{Jinan3}} & \multicolumn{2}{c}{\textbf{Hangzhou1}} & \multicolumn{2}{c}{\textbf{Hangzhou2}} & \multirow{2}{*}{\textbf{Paradigm}} \\ \cline{2-11}
 & ATT$\downarrow$ & AWT$\downarrow$ & ATT$\downarrow$ & AWT$\downarrow$ & ATT$\downarrow$ & AWT$\downarrow$ & ATT$\downarrow$ & AWT$\downarrow$ & ATT$\downarrow$ & AWT$\downarrow$ &  \\ \hline
FixedTime & 453.41 & 51.32 & 370.34 & 35.15 & 384.53 & 36.95 & 497.54 & 36.41 & 408.53 & 53.94 & \multirow{2}{*}{\textit{\begin{tabular}[c]{@{}c@{}}Traditional\\ Methods\end{tabular}}} \\
Maxpressure & 274.34 & 32.04 & 246.35 & 22.56 & 245.66 & 24.31 & 289.55 & 21.52 & 349.53 & 67.52 &  \\ \hline
LLMLight-7B [KDD'24] & 274.47 & 33.66 & 286.53 & 28.66 & 271.11 & 28.27 & 299.31 & 25.53 & 331.38 & 51.79 & \multirow{6}{*}{\textit{\begin{tabular}[c]{@{}c@{}}RL-\\ based\\ Methods\end{tabular}}} \\
MPLight [AAAI'20] & 455.34 & 72.45 & 471.14 & 78.03 & 427.37 & 64.91 & 491.32 & 64.05 & 425.42 & 69.85 &  \\
AttendLight [NeurIPS'20] & 381.11 & 67.59 & 305.53 & 64.72 & 331.34 & 66.42 & 318.94 & 67.84 & 348.41 & 65.58 &  \\
CoLight [CIKM'19] & 472.44 & 91.09 & 450.41 & 78.59 & 498.84 & 89.94 & 494.61 & 72.18 & 435.32 & 81.11 &  \\
Efficient-Colight [Arxiv'21] & 663.16 & 98.98 & 640.34 & 91.32 & 638.23 & 80.34 & 701.45 & 103.43 & 534.94 & 87.19 &  \\
Advanced-CoLight [ICML'22] & 347.31 & 56.54 & 345.78 & 35.96 & 342.56 & 37.55 & 485.32 & 54.11 & 523.19 & 72.56 &  \\ \hline
CoLLMLight-8B [Arxiv'25] & 281.12 & 33.23 & 269.34 & 25.51 & 268.32 & 34.36 & 298.42 & 24.45 & 336.92 & 45.43 & \multirow{8}{*}{\textit{\begin{tabular}[c]{@{}c@{}}Zero-\\ shot\\ Methods\end{tabular}}} \\
Llama3.3-70B [Meta'24] & {\ul 272.41} & 33.53 & {\ul 244.55} & 22.04 & {\ul 243.53} & 25.43 & 281.44 & 17.65 & 326.42 & 45.56 &  \\
Qwen 2.5-72B [Alibaba'24] & 275.42 & 33.15 & 251.41 & 25.49 & 264.21 & 24.54 & 282.13 & 17.54 & 329.34 & 39.34 &  \\
GPT 3.5-turbo [OpenAI'23] & 337.32 & 39.98 & 328.19 & 37.08 & 343.19 & 34.35 & 293.42 & 23.45 & 348.59 & 33.45 &  \\
GPT-4o [OpenAI'24] & 281.58 & {\ul 30.11} & 259.61 & 24.71 & 258.85 & {\ul 24.17} & 280.48 & {\ul 16.32} & {\ul 325.48} & 32.26 &  \\
DeepSeek-R1-671B [DeepSeek'25] & 279.11 & 31.85 & 258.43 & {\ul 21.67} & 262.21 & 27.87 & {\ul 278.565} & 17.81 & \multicolumn{1}{l}{335.53} & \multicolumn{1}{l}{\textbf{30.19}} &  \\
DeepSeek-R1-Distill-7B [DeepSeek'25] & 331.45 & 38.91 & 311.43 & 31.43 & 288.42 & 29.23 & 291.32 & 19.56 & 344.73 & 33.72 &  \\
\textbf{Traffic-R1-3B (Ours)} & \textbf{270.34} & \textbf{27.95} & \textbf{239.53} & \textbf{21.11} & \textbf{238.03} & \textbf{23.17} & \textbf{277.83} & \textbf{15.51} & \textbf{324.11} & {\ul 33.14} &  \\ \hline
\end{tabular}%
}
\vspace{-1.5em}
\end{table*}
\label{sec:eval}
In this section, we evaluate our proposed \model to address the following research questions:
\vspace{-1em}
\begin{itemize}[leftmargin=*]
    \item \textbf{RQ1}: Can \model outperform other TSC expert models and LLMs on public datasets and in zero-shot settings?
    \vspace{-1em}
    %here we will present the big table like LLM light for represent exp.
    \item \textbf{RQ2}: How does \model perform in handling OOD incidents through its reasoning?
    %here will provide a performance circle
    \vspace{-1em}
    \item \textbf{RQ3}: What's the advantage of our RL-based finetuning over traditional paradigms for LLMs in traffic control tasks?
    \vspace{-1em}
    \item \textbf{RQ4}: How effective are the designs of \model under various ablation settings?
\end{itemize}
\vspace{-1em}

\subsection{Experimental Settings}

\subsubsection{Dataset}
\vspace{-0.5em}
Our experiments were primarily conducted on two public traffic flow datasets~\cite{mei2024libsignal} to ensure fair comparison, as detailed in Table \ref{tab:dataset}.
For out-of-distribution scenarios, we collect traffic emergency incident recordings from traffic management departments and summarize them into 200 representative textual examples, such as passages running onto roads, vehicle accidents, and school times, along with the action records implemented by traffic managers as correct responses (presented in Appendix \ref{app:incident}). Besides, for , we modify the Hangzhou1 datasets by incorporating a 5\% proportion of emergency vehicle flow to simulate emergency vehicle coordination scenarios.

\subsubsection{Implementation Details}
%\vspace{-0.5em}
\model was trained based on Qwen2.5-3B-base on a device with 4 H100 GPUs and deployed for inference on a single Tesla T40 GPU, utilizing the Verl framework~\cite{sheng2024hybridflow}. We evaluated all models using the CityFlow traffic simulator~\cite{zhang2019cityflow}.
The experimental setup features a standard four-phase action space: north-south through (NTST), east-west through (ETWT), east-west left-turn (ELWL), and north-south left-turn (NLSL). Each green phase lasts 15 seconds, followed by a 3-second yellow and a 2-second red transition period, consistent with real-world TSC systems~\cite{zhang2022expression,mei2024libsignal}. Right turns are permitted at all times. All traffic flow datasets are simulated for one-hour periods.

\subsubsection{Baseline Methods} 
%\vspace{-0.5em}
We incorporate a range of baseline models from various research areas to ensure a comprehensive comparison. For traditional TSC methods, we include FixedTime~\cite{koonce2008traffic} and Maxpressure~\cite{varaiya2013max}. For RL-based methods, we evaluate five effective approaches: MPLight~\cite{chen2020toward}, AttendLight~\cite{oroojlooy2020attendlight}, CoLight~\cite{wei2019colight}, Efficient-CoLight~\cite{wu2021efficient}, and Advanced-CoLight~\cite{zhang2022expression}, along with the state-of-the-art LLM-based method, LLMLight~\cite{lai2023llmlight}.
For zero-shot methods, we assess the performance of CoLLMLight~\cite{yuan2025collmlight} and general LLM models, which include Llama 3.3 (70B), Qwen 2.5 (72B), GPT 3.5-turbo, GPT-4o, and DeepSeek-R1 (671B and distilled to 7B). All learning-based baselines are trained on the same 4x4 simulated road network and traffic flow dataset as \model during the open-world RL stage. Notably, for LLMLight and CoLLMLight, we also incorporate our expert-collaborative dataset into the training instructions to ensure a fair comparison
\vspace{-0.5em}
\subsubsection{Evaluation Protocols}
\vspace{-0.5em}
We adopt the commonly used Average Travel Time (ATT) and Average Waiting Time (AWT) to evaluate the performance of models. Lower values in ATT and AWT indicate better traffic efficiency. 

\vspace{-0.5em}
\subsection{Conventional TSC Evaluation (RQ1)}
\vspace{-0.5em}
We evaluate the performance of \model on conventional TSC tasks using public datasets that are widely adopted in TSC research. All learning-based methods are trained in the same simulated traffic environment as \model to fairly assessment. As shown in Table~\ref{tab:exp}, our model significantly outperforms all baselines, demonstrating strong generalization for real-world deployment.
RL-based methods perform poorly in zero-shot scenarios, even lagging behind traditional methods, which questions their real-world applicability without iterative training. While large LLMs like DeepSeek-R1-671B achieve impressive zero-shot results, their performance degrades when distilled into smaller models.
For completeness, Appendix~\ref{app:moreexp} presents full-shot results, where RL models are trained on each test dataset. Although RL methods then surpass traditional ones, our zero-shot model still outperforms them, highlighting its superior internal reasoning and control policies.
% We evaluate the performance of \model on standard signal control tasks, which are widely studied in most TSC researches. We adopt the original deployment settings of all learning-based baseline models for training on the test environments, while \model is assessed directly in a zero-shot setting. The results, presented in Table~\ref{tab:exp}, show that \model, in its zero-shot configuration, outperforms all baselines, including RL-based methods that underwent extensive training and optimization on the test datasets. Notably, some advancing LLMs achieve impressive performance comparable to SOTA learning-based methods, demonstrating the potentiality of LLMs for taking place of humans on traffic tasks. Moreover, \model achieve better performance than SOTA LLMs (e.g. DeepSeek-R1-671B) with only 1\% of their parameter size.

\subsection{OOD Incident Dealing (RQ2)}
Evaluation only in ideal simulations is insufficient for deployment. RQ2 asks whether \model can handle out-of-distribution (OOD) traffic incidents via human-like reasoning. Incident handling requires internal logic, commonsense, and domain knowledge, which challenge both traditional RL and standard LLM approaches.
% \usepackage{multirow}
% \usepackage{graphicx}
% \usepackage[normalem]{ulem}

\begin{table}[b]
\vspace{-1em}
\centering
\footnotesize
\caption{OOD Incident Dealing Evaluation. ``-'' indicates the method is entirely inadequate for the task.}
\label{tab:location}
\vspace{-1em}
\resizebox{\columnwidth}{!}{%
\begin{tabular}{ccccc}
\hline
\multirow{2}{*}{Method} & \multirow{2}{*}{Parameter Size} & Local & \multicolumn{2}{c}{Network-wide} \\ \cline{3-5} 
 &  & EAA$\uparrow$ & AETT$\downarrow$ & AEWT$\downarrow$ \\ \hline
Random & n.a. & 0.25 & 614.45 & 97.42 \\
MaxPressure & n.a. & - & 287.94 & 21.87 \\
Advanced-CoLight & n.a. & - & 286.32 & 24.53 \\
LALight & 72B & 0.82 & 234.42 & 12.32 \\
LLMLight & 7B & 0.42 & 273.55 &15.21  \\
Qwen2.5 (large) & 72B & 0.88  & 232.54 & 10.53 \\
DeepSeek-R1 & 672B & \textbf{0.93} & {\ul 223.19} & {\ul 10.14} \\ \hline
Traffic-R1 & 3B & {\ul 0.85} & \textbf{215.58} & \textbf{7.98} \\ \hline
% \multicolumn{5}{r}{("-" indicates the method is totally incapable for the task)}\\
\end{tabular}%
}
\vspace{-1em}
\end{table}
We split OOD tasks into two types: \textbf{local intersection incidents} (single-intersection events, e.g., an local accident) and \textbf{network-wide incidents} (events affecting multiple intersections, e.g., emergency-vehicle routing). For local incidents we use Emergency Action Accuracy (EAA). For network-wide incidents we use Average Emergency Travel Time (AETT) and Average Emergency Waiting Time (AEWT) adapted from \citet{wang2024llm}. Results are reported in Table~\ref{tab:location}. The findings highlight two key strengths of \model:
\vspace{-0.5em}
\begin{itemize}[leftmargin=*]
\item \textbf{Stable OOD generalization.} \model maintains consistent performance across diverse OOD scenarios and outperforms larger general LLMs and LLMLight by over 30\% across metrics, indicating it applies traffic knowledge instead of merely imitating training examples.
\vspace{-0.5em}
\item \textbf{Lightweight and efficient.} At 3B parameters, \model matches or exceeds traditional baselines and advanced LLMs while using roughly 1\% of their parameters and much lower deployment cost. It performs particularly well on network-wide incidents, showing that the asynchronous communication design enables coordinated multi-intersection control.
\end{itemize}
\vspace{-1em}

\subsection{Discussion of RL-based Finetuning (RQ3)}
\vspace{-0.5em}
To validate our proposed RL-based two-stage finetuning paradigm, we compare it against traditional instruction finetuning, represented by the LLMLight framework~\cite{lai2023llmlight}. For a fair comparison, both methods use the same Qwen2.5 base model. We built the instruction dataset for LLMLight using a similar pipeline: RL-generated action trajectories, GPT-4 explanations, and expert-collaborative QA samples. After training, we evaluate the models in two dimensions: \textit{1. TSC performance:} we assess zero-shot capabilities, quantifying results based on the average improvement in ATT and AWT across all datasets, with Qwen2.5-3B's performance normalized to 0.5. \textit{2. General capabilities:} we use public benchmarks~\cite{evalscope_2024} to evaluate reasoning, instruction following, and commonsense to gauge the models' stability and potential for real-world deployment.
The results in Figure~\ref{fig:RQ3}, highlight two key findings regarding RL finetuning for LLMs in traffic tasks:
\begin{figure}[b]
    \centering
    \vspace{-1em}
    \includegraphics[width=1.0\linewidth]{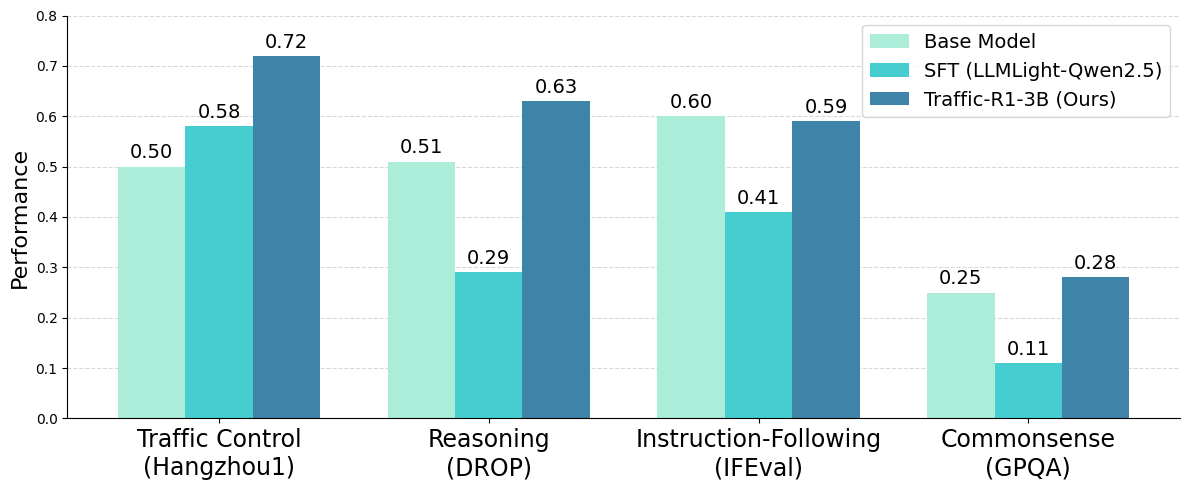}
    \vspace{-2.5em}
    \caption{Comparison results for models' capacities.}
    \vspace{-1em}
    \label{fig:RQ3}
\end{figure}

\vspace{-1em}
\begin{itemize}[leftmargin=*]
\item \textbf{From imitation to reasoning.} \model significantly outperforms the instruction-finetuned model on zero-shot TSC tasks. This performance gap, despite identical training data, shows that instruction finetuning encourages imitation of synthetic data. In contrast, our RL-based paradigm fosters internal reasoning through self-exploration and iteration.
\vspace{-1em}
\item \textbf{Complete general capabilities.} Instruction finetuning often causes "catastrophic forgetting," degrading LLM's general abilities. The SFT model exemplifies this, performing worse than the base model on general benchmarks. \model mitigates this issue using self-rollout samples and KL divergence penalty. This approach constrains the policy update to a controlled space, promoting deeper optimization over memorization and preserving the model's comprehensive capabilities.
\end{itemize}

\subsection{Ablation Study (RQ4)}

\begin{figure}[!t]
    \centering
    \vspace{-1em}
    \includegraphics[width=1\linewidth]{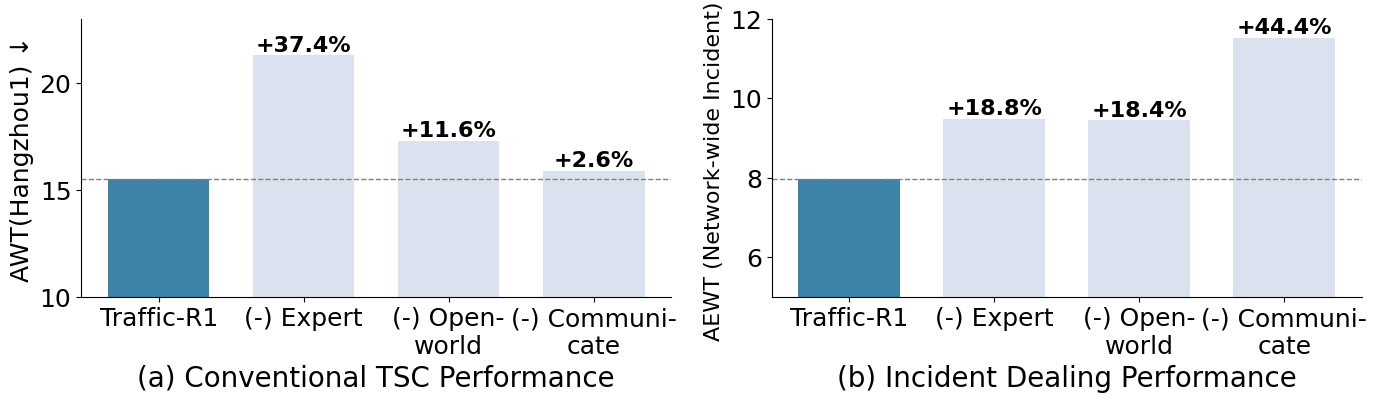}
    \vspace{-2em}
    \caption{Ablation results on TSC and OOD tasks.}
    \vspace{-1.5em}
    \label{fig:ablation}
\end{figure}

To assess the contribution of each component of \model to its performance, we developed the following model variants for our ablation study:
\vspace{-0.5em}
\begin{itemize}[leftmargin=*]
    \item \textbf{(-) Expert.} This variant excludes the human-informed RL stage and is trained solely using open-world exploration.
    \vspace{-0.5em}
    \item \textbf{(-) Open-world.} This variant excludes the open-world RL stage during training.
    \vspace{-0.5em}
    \item \textbf{(-) Communicate.} This variant removes the asynchronous communication mechanism and operates without communication.
\end{itemize}
\vspace{-1em}
We present the ablation results for \model and its variants on both conventional traffic scenarios and OOD scenarios in Figure~\ref{fig:ablation}. Our findings are summarized as follows: 1) The human-informed RL stage is necessary to establish foundational TSC knowledge, enabling stable exploration in the subsequent open-world RL stage. 2) The open-world RL stage is effective in unlocking the model's potential to achieve superior performance. 3) Asynchronous communication is critical for multi-intersection coordination tasks but is not determinative for conventional traffic scenarios.

\vspace{-0.5em}

\section{Real-world deployment}
\vspace{-0.5em}

\begin{figure}[h]
\centering
\vspace{-1em}
\includegraphics[width=1\linewidth]{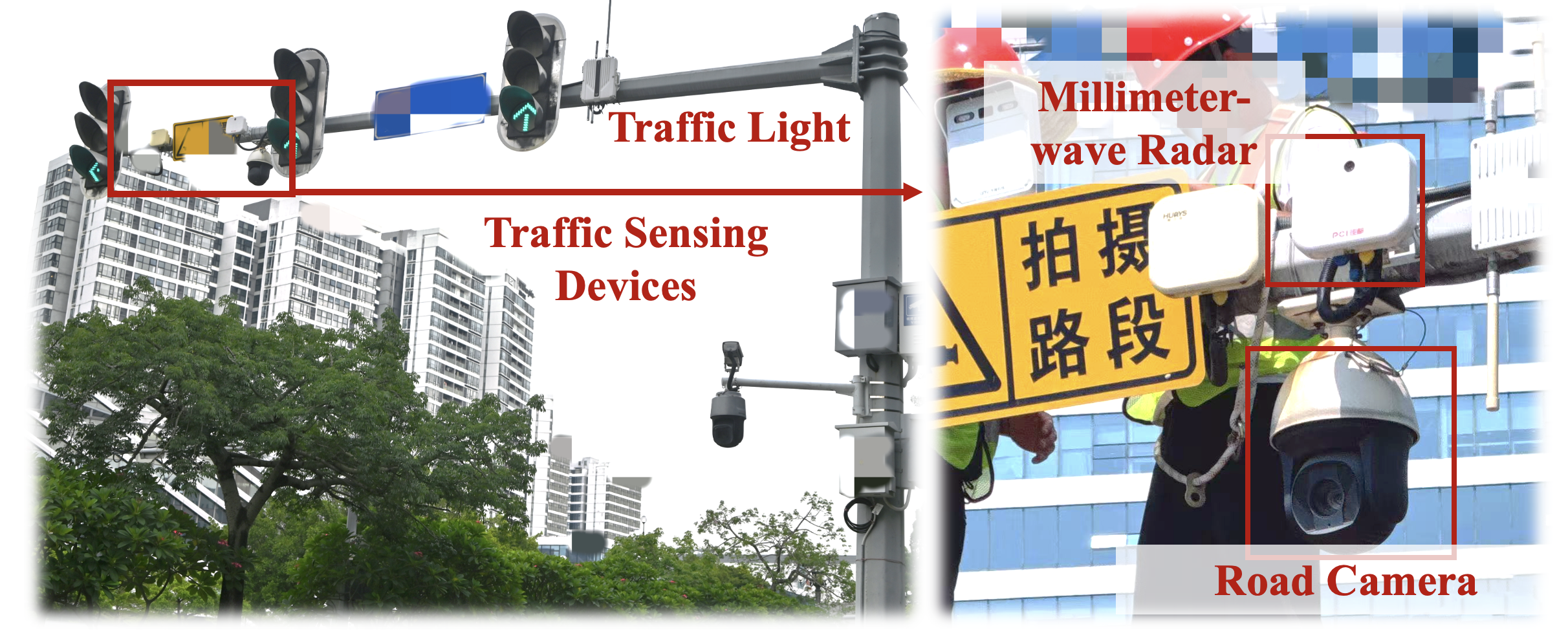}
\vspace{-2em}
\caption{In-suit traffic sensing devices for Traffic-R1}
\vspace{-1em}
\label{fig:device}
\end{figure}
As shown in Figure \ref{fig:device}, \model is deployed on a online traffic platform in a Chinese city (name withheld for privacy), controlling 10 key intersections and serving over 55,000 drivers daily. A/B tests for 6 weeks (Table~\ref{tab:deploy}) show significant improvements in real-world traffic control—more efficient decision-making, superior performance compared with human experts, and consistent operation with reduced labor.
We place details and discussion in Appendix \ref{app:deploy} due to space limit.
\begin{table}[!h]
\centering
\vspace{-0.5em}
\caption{A/B test results spanning 6 weeks.}
\vspace{-1em}
\label{tab:deploy}
\resizebox{\columnwidth}{!}{%
\begin{tabular}{cccc}
\hline
\textbf{Method} & Average Queue$\downarrow$ & Maximum Queue$\downarrow$ & Working Hours$\downarrow$ \\ \hline
Manual & 34.5 & 50.3 & 2+ \\
Traffic-R1 & 31.3 & 48.1 & 0.5+ \\
\textbf{\#Improve} & \textbf{9.3\%} & \textbf{4.4\%} & \textbf{$\sim$75\%} \\ \hline
\end{tabular}%
}
\vspace{-1em}
\end{table}

\vspace{-0.5em}
\section{Conclusion and Future Work}
\vspace{-0.3em}
\label{sec:conclusion}
In this paper, we introduce a two-stage reinforcement learning strategy and a network communication framework to convert the LLM into a foundational traffic-control model that operates like a human traffic agent. Evaluation on standard TSC benchmarks and traffic-incident handling shows improvement over prior methods, and real-world tests demonstrate value for industrial deployment. Future work includes reinforced VLMs for direct road-vision reasoning without reliance on structured textual traffic data, enhancing efficiency and deployment convenience. 

\clearpage
\section{Limitations}
While \model demonstrates strong performance in both standard and out-of-distribution traffic control tasks with impressive zero-shot generalization, it has several limitations. First, the base model, training, and inference are highly reliant on structured data in a single textual modality: traffic observations and incident information must be converted into structured text before being input to the model. This requirement causes inconvenience and potential information loss in dense real-world deployments, and we will address it in future work by developing reinforced vision-language models and multimodal input pipelines. Second, due to limitations in deployment scale and practical constraints imposed by traffic management agencies, we cannot provide a detailed analysis of LLM hallucination-related instability across large-scale traffic scenarios or assess drivers' feedback to the agent-based traffic network. Access restrictions, privacy requirements, safety constraints, and the cost and complexity of long-term field trials prevented extended real-world evaluations and systematic collection of driver response data.

\bibliography{acl_ref}

\clearpage
\appendix 
\section{Appendix}
% \balance
\label{sec:appendix}
\subsection{Input Template for Traffic-R1}
\label{app:prompt}
\begin{cmt*}{Input Templates for LLM Rollout}{}
%\vspace{-1em}
\textbf{System:} You are a helpful traffic control agent.
% \vspace{10pt}
\\
\textbf{Task Description: }The crossroad connects two roads: north-south and east-west, with the traffic light at their intersection. Each road is divided into two sections (e.g., north and south for the north-south road) and each section has two lanes: a through lane and a left-turn lane...\\
\textbf{Structured Traffic Observation: }\\
Signal: ETWT\\
Allowed lanes: Eastern and western through lanes\\
- Early queued: 2 (East), 1 (West), 3 (Total)\\
- Segment 1: 0 (East), 0 (West), 0 (Total)\\
- Segment 2: 1 (East), 0 (West), 1 (Total)\\
- Segment 3: 1 (East), 2 (West), 3 (Total)\\
\vspace{2pt}
Signal: ELWL\\
Allowed lanes: Eastern and western left lanes\\
- Early queued: 0 (East), 1 (West), 1 (Total)\\
- Segment 1: 0 (East), 0 (West), 0 (Total)\\
- Segment 2: 3 (East), 0 (West), 3 (Total)\\
- Segment 3: 1 (East), 0 (West), 1 (Total)\\
\vspace{2pt}
Signal: NTST\\
Allowed lanes: North and south through lanes\\
- Early queued: 0 (North), 0 (South), 0 (Total)\\
- Segment 1: 0 (North), 0 (South), 0 (Total)\\
- Segment 2: 1 (North), 0 (South), 1 (Total)\\
- Segment 3: 1 (East), 1 (West), 2 (Total)\\
\vspace{2pt}
Signal: NLSL\\
Allowed lanes: North and south left lanes\\
- Early queued: 1 (North), 0 (South), 1 (Total)\\
- Segment 1: 0 (North), 0 (South), 0 (Total)\\
- Segment 2: 0 (North), 0 (South),0 (Total)\\
- Segment 3: 1 (East), 0 (West), 1 (Total)\\
\vspace{2pt}
\textbf{Incident Information (optional):}\\
(at training stage we do not use this part)\\
Refer Appendix \ref{app:incident}...\\
\textbf{Format Instruction: }You can only choose one of the signals listed above. 
You FIRST think about the reasoning process for your choice as an internal monologue and then provide the final answer. 
Your think process MUST BE put in <think>...</think> tags. The final choice MUST BE put in \textbackslash{}boxed\{ \}.
\vspace{-1em}
\end{cmt*}

% \usepackage{multirow}
% \usepackage{graphicx}
% \usepackage[normalem]{ulem}
\begin{table*}[]
\centering
\caption{Performance comparison on conventional traffic signal control tasks (the smaller the better). The best results are in \textbf{bold} and second-best results are {\ul underlined}.}
\vspace{-1em}
\label{tab:app}
\resizebox{1.0\textwidth}{!}{%
\begin{tabular}{cccccccccccc}
\hline
\multirow{2}{*}{\textbf{Models}} & \multicolumn{2}{c}{\textbf{Jinan1}} & \multicolumn{2}{c}{\textbf{Jinan2}} & \multicolumn{2}{c}{\textbf{Jinan3}} & \multicolumn{2}{c}{\textbf{Hangzhou1}} & \multicolumn{2}{c}{\textbf{Hangzhou2}} & \multirow{2}{*}{\textbf{Paradigm}} \\ \cline{2-11}
 & ATT & AWT & ATT & AWT & ATT & AWT & ATT & AWT & ATT & AWT &  \\ \hline
FixedTime & 453.41 & 51.32 & 370.34 & 35.15 & 384.53 & 36.95 & 497.54 & 36.41 & 408.53 & 53.94 & \multirow{2}{*}{\textit{\begin{tabular}[c]{@{}c@{}}Traditional\\ Methods\end{tabular}}} \\
Maxpressure & 274.34 & 32.04 & 246.35 & 22.56 & 245.66 & 24.31 & 289.55 & 21.52 & 349.53 & 67.52 &  \\ \hline
LLMLight-7B & 274.47 & 33.66 & 256.53 & 28.66 & 247.11 & 28.27 & 289.31 & 25.53 & 331.38 & 51.79 & \multirow{6}{*}{\textit{\begin{tabular}[c]{@{}c@{}}RL-\\ based\\ Methods\end{tabular}}} \\
MPLight & 310.54 & 50.45 & 270.14 & 48.03 & 272.37 & 42.91 & 319.32 & 44.05 & 365.42 & 69.85 &  \\
AttendLight & 280.11 & 47.59 & 250.53 & 34.72 & 251.34 & 36.42 & 288.94 & 27.84 & 338.41 & 55.58 &  \\
CoLight & 272.44 & 41.09 & 250.41 & 38.59 & 248.84 & 39.94 & 294.61 & 42.18 & 335.32 & 61.11 &  \\
Efficient-Colight & {\ul 263.16} & 28.98 & 240.34 & {\ul 21.32} & {\ul 238.23} & \textbf{20.34} & 301.45 & 33.43 & 334.94 & 47.19 &  \\
Advanced-CoLight & \textbf{247.31} & 32.54 & \textbf{235.78} & 25.96 & 242.56 & 27.55 & 285.32 & 24.11 & 323.19 & 52.56 &  \\ \hline
Llama3.3-70B & 272.41 & 33.53 & 244.55 & 22.04 & 243.53 & 25.43 & 281.44 & 17.65 & 326.42 & 45.56 & \multirow{7}{*}{\textit{\begin{tabular}[c]{@{}c@{}}Zero-\\ shot\\ Methods\end{tabular}}} \\
Qwen 2.5-72B & 275.42 & 33.15 & 251.41 & 25.49 & 264.21 & 24.54 & 282.13 & 17.54 & 329.34 & 39.34 &  \\
GPT 3.5-turbo & 337.32 & 39.98 & 328.19 & 37.08 & 343.19 & 34.35 & 293.42 & 23.45 & 348.59 & 33.45 &  \\
GPT-4o & 281.58 & {\ul 30.11} & 259.61 & 24.71 & 258.85 & 24.17 & 280.48 & {\ul 16.32} & {\ul 325.48} & {\ul 32.26} &  \\
DeepSeek-R1-671B & 279.11 & 31.85 & 258.43 & 21.67 & 262.21 & 27.87 & {\ul 278.565} & 17.81 & \multicolumn{1}{l}{335.53} & \multicolumn{1}{l}{\textbf{30.19}} &  \\
\multicolumn{1}{l}{DeepSeek-R1-Distill-7B} & \multicolumn{1}{l}{331.45} & \multicolumn{1}{l}{38.91} & \multicolumn{1}{l}{311.43} & \multicolumn{1}{l}{31.43} & \multicolumn{1}{l}{288.42} & \multicolumn{1}{l}{29.23} & \multicolumn{1}{l}{291.32} & \multicolumn{1}{l}{19.56} & \multicolumn{1}{l}{344.73} & \multicolumn{1}{l}{33.72} &  \\
\textbf{Traffic-R1-3B} & 270.34 & \textbf{27.95} & {\ul 239.53} & \textbf{21.11} & \textbf{238.03} & {\ul 23.17} & \textbf{277.83} & \textbf{15.51} & \textbf{324.11} & 33.14 &  \\ \hline
\end{tabular}%
}
\vspace{-1em}
\end{table*}
\subsection{Introduction of baselines}
We compare our method against three categories of approaches for traffic control. Below is detailed information about these methods:
\begin{itemize}[leftmargin=0pt]
    \item \textbf{Traditional Methods:} This category includes conventional traffic signal control (TSC) methods, which are straightforward and widely adopted in real-world traffic systems.
    \begin{itemize}
        \item \textbf{FixedTime~\cite{koonce2008traffic}:} A policy that assigns a fixed cycle length with predefined phase splits across all phases.
        \item \textbf{MaxPressure~\cite{varaiya2013max}:} A control strategy that selects the phase with the highest pressure to optimize traffic flow.
    \end{itemize}
    \item \textbf{RL-based Methods:} These methods normally require training and interaction with their policies on each evaluation dataset.
    \begin{itemize}
        \item \textbf{LLMLight-7B~\cite{lai2023llmlight}:} A SOTA LLM-based TSC method that employs the Advanced-CoLight framework to interact and generate action policies for each dataset. It utilizes GPT-4 to generate explanations for each action, which, along with Advanced-CoLight generated actions, are used for instruction finetuning to enable the LLM base model to emulate TSC capabilities.
        \item \textbf{MPLight~\cite{chen2020toward}:} A method based on the FRAP model that uses pressure as both observation and reward to optimize TSC.
        \item \textbf{AttendLight~\cite{oroojlooy2020attendlight}:} A method that employs attention mechanism to construct phase features and predict its transition probabilities.
        \item \textbf{CoLight~\cite{wei2019colight}:} A method that uses a graph attention network to represent inter-intersection communication within a RL framework.
        \item \textbf{Efficient-CoLight~\cite{wu2021efficient}:} An enhanced version of the CoLight model that incorporates efficient pressure as an observation to improve decision-making in TSC.
        \item \textbf{Advanced-CoLight~\cite{zhang2022expression}:} A SOTA RL-based method that enhances CoLight by integrating efficient pressure and advanced traffic state features, such as effective running vehicles, to optimize decision-making capabilities.
    \end{itemize}
    \item \textbf{Zero-shot Methods:} These methods (including \model) are represented by their zero-shot working ability on every dataset. In our experiment, the models are not trained on any TSC evaluation dataset and carry out traffic control based on their inner policies and knowledge.
    \begin{itemize}
    \item \textbf{CoLLMLight-8B~\cite{yuan2025collmlight}:} An enhanced version of LLMLight~\cite{lai2023llmlight} that incorporates neighbor information into consideration. The model is trained using an instruction fine-tuning framework on a synthetic dataset and demonstrates improved zero-shot performance compared to LLMLight.
    \item \textbf{Llama3.3-70B:} A LLM developed by Meta AI, featuring 70 billion parameters and a 128K token context window. It offers performance comparable to much larger models in zero-shot tasks, with enhanced capabilities in tool calling and multilingual support.
    \item \textbf{Qwen2.5-72B:} A LLM developed by Alibaba Cloud, with 72 billion parameters. It is designed for superior performance in zero-shot learning, particularly in coding, mathematics, and following complex instructions.
    \item \textbf{GPT-3.5-turbo:} A variant of OpenAI's GPT-3.5 model, finetuned for conversational applications. It excels in zero-shot dialogue systems and natural language processing tasks.
    \item \textbf{GPT-4o:} OpenAI's advancing multimodal language model that processes text, audio, and images. It provides advanced zero-shot capabilities across multiple tasks, with fast response times and cost efficiency.
    \item \textbf{DeepSeek-R1-671B:} A massive language model with 671 billion parameters, developed by DeepSeek AI. It specializes in zero-shot reasoning, particularly in mathematics and coding, and is available as an open-source model.
    \item \textbf{DeepSeek-R1-Distill-7B:} A Qwen2.5-based distilled version of DeepSeek-R1, featuring 7 billion parameters. It maintains competitive zero-shot performance in reasoning tasks through efficient distillation methods.
    \end{itemize}
\end{itemize}

\begin{figure*}[!h]
\begin{cmt*}{Responses of \model in different scenarios}{}
\textbf{Case 1: }<think>Based on the current state of the intersection, Signal: ETWT (Eastern and Western through lanes) is the one that most significantly improves traffic conditions during the next phase. This is because this lane has \textcolor{tblue}{\textbf{18 early queued vehicles and 13 approaching vehicles in Segment 3}}, indicating a significant number of vehicles waiting to pass through and \textcolor{tblue}{\textbf{incoming huge traffic pressure}}. Approving this signal will allow these vehicles to proceed, easing the traffic flow.</think>\\

\textbf{Case 2: }<think>Based on the current state of the intersection, the signal phase that will most significantly improve traffic conditions during the next phase is NTST (Northern and southern through lanes). This is because \textcolor{tblue}{\textbf{only the southern through lane has approaching vehicles in Segment 1}}, which indicates that there are vehicles waiting to enter the intersection. By allowing vehicles in the northern and southern through lanes to flow, \textcolor{tblue}{\textbf{this signal phase will help avoid the queue of approaching vehicles}} and improve overall traffic flow.</think>\\

\textbf{Case 3: }<think>Based on the current state of the intersection, the signal phase that will most significantly improve traffic conditions during the next phase is ETWT (Eastern and Western through lanes). Although the \textcolor{tblue}{\textbf{northern through lane has 20 early queued vehicles in Segment 2}}, an ambulance is currently approaching from the east, heading west. Prioritizing ETWT is critical to ensure \textcolor{tblue}{\textbf{the ambulance passes through the intersection without delay}}, maintaining emergency response efficiency. This signal will also alleviate some traffic pressure by allowing eastern and western vehicles to proceed.</think>\\

\textbf{Case 4: }<think>Based on the current state of the intersection, the signal phase that will most significantly improve traffic conditions during the next phase is NTST (Northern and Southern pedestrian crossing). This is because \textcolor{tblue}{\textbf{a large number of students are waiting to cross from the northern to southern direction}}, indicating significant pedestrian demand. Activating this signal will allow these students to cross safely, reducing congestion at the crosswalk.</think>\\

\textbf{Case 5: }<think>Based on the current state of the intersection and the urgent message of an emergency fire truck approaching from the west, the signal phase that will most significantly improve traffic conditions during the next phase is ETWT (Eastern and Western through lanes). This is because \textcolor{tblue}{\textbf{both ETWT and NTST have comparable queues, with approximately 15 early queued vehicles each in Segment 2, but prioritizing ETWT ensures the fire truck's swift passage}}. Activating this signal will clear the western lane, facilitating emergency access while also easing general traffic flow.</think>\\
\end{cmt*}
\vspace{-2em}
\end{figure*}

\subsection{Traffic Incident Data}
\label{app:incident}

In this work, we evaluate the models' incident handling capabilities by collecting traffic emergency incident recordings from traffic management departments. We summarize these recordings into 200 representative textual examples to simulate the occurrence of out-of-distribution (OOD) traffic incidents. Below, we present selected anonymized examples of these incident recordings:

\begin{cmt*}{Examples for testing traffic incidents}{}
%\vspace{-1em}
\textbf{ID 1:} At this intersection, a traffic accident in the eastbound lane is causing significant congestion.
\textbf{Response Action:} \textcolor{lightgreen}{ETWT}
\vspace{2pt}
\\
\textbf{ID 2:} At this intersection, a pedestrian was struck in the northbound crosswalk.
\textbf{Response Action:} \textcolor{lightgreen}{ETWT/NLSL}
\vspace{2pt}
\\
\textbf{ID 3:} Report from the nearby intersection to the north: Heavy southbound traffic is approaching.
\textbf{Response Action:} \textcolor{lightgreen}{NTST/NLSL}
\vspace{2pt}
\\
\textbf{ID 4:} At this intersection, a school bus is stopped in the eastbound lane, loading students.
\textbf{Response Action:} \textcolor{lightgreen}{NTST}
\vspace{2pt}
\\
\textbf{ID 5:} At this intersection, a group of pedestrians is blocking the westbound crosswalk.
\textbf{Response Action:} \textcolor{lightgreen}{NTST}
\vspace{2pt}
\\
\textbf{ID 6:} Report from the nearby intersection to the east: A fire hydrant crew is slowing westbound traffic.
\textbf{Response Action:} \textcolor{lightgreen}{NTST/ELWL}
\vspace{2pt}
\\
\textbf{ID 7:} At this intersection, vehicles spun out in the westbound lane.
\textbf{Response Action:} \textcolor{lightgreen}{NTST}
\vspace{2pt}
\\
\textbf{ID 8:} At this intersection, for the XXX event, a marathon is passing through the eastbound lane.
\textbf{Response Action:} \textcolor{lightgreen}{ETWT}
\vspace{2pt}
\\
\textbf{ID 9:} At this intersection, a road rage incident was reported in the northbound lane due to a heavy traffic jam.
\textbf{Response Action:} \textcolor{lightgreen}{NTST}
\vspace{2pt}
\\
\textbf{ID 10:} At this intersection, it is 5:30 PM, the nearby western school is scheduled to dismiss, leading to increased vehicular traffic for student pick-up.
\textbf{Response Action:} \textcolor{lightgreen}{ETWT}
\\

\vspace{-1em}
\end{cmt*}

\subsection{More Experiment Result}
\label{app:moreexp}
We evaluate the performance of \model on standard signal control tasks in full-shot settings. We adopt the original deployment configurations of all learning-based baseline models for training in the test environments, while \model is assessed directly in a zero-shot setting. The results, presented in Table~\ref{tab:app}, show that \model, in its zero-shot configuration, outperforms all baselines, including RL-based methods that underwent extensive training and optimization on the test datasets. Notably, some advanced LLMs achieve impressive performance comparable to SOTA RL-based methods, demonstrating the potential of LLMs to fully replace traditional RL-based methods in traffic control tasks. Furthermore, \model achieves better performance than SOTA LLMs (e.g., DeepSeek-R1-671B) while utilizing only 1\% of their parameter size.

\subsection{Reasoning Cases of \model}
\label{app:case}

We provide several representative reasoning processes generated by \model during experiments across different scenarios. The results clearly demonstrate the model's comprehensive and logical reasoning ability in both conventional traffic situations and incident response tasks. The reasoning examples illustrate the model's understanding of traffic control, including its focus on queued vehicles and approaching vehicles. The model independently makes trade-offs to achieve better traffic efficiency.
Furthermore, when an incident occurs, \model accurately analyzes the situation and identifies optimal actions to address it, while even striving to both maximize traffic efficiency and respond to the incident's needs. This showcases the model's ability to balance multiple objectives in dynamic traffic environments.

\subsection{Deployment Details}
\label{app:deploy}

\begin{figure}[b]
\centering
\vspace{-1em}
\includegraphics[width=1\linewidth]{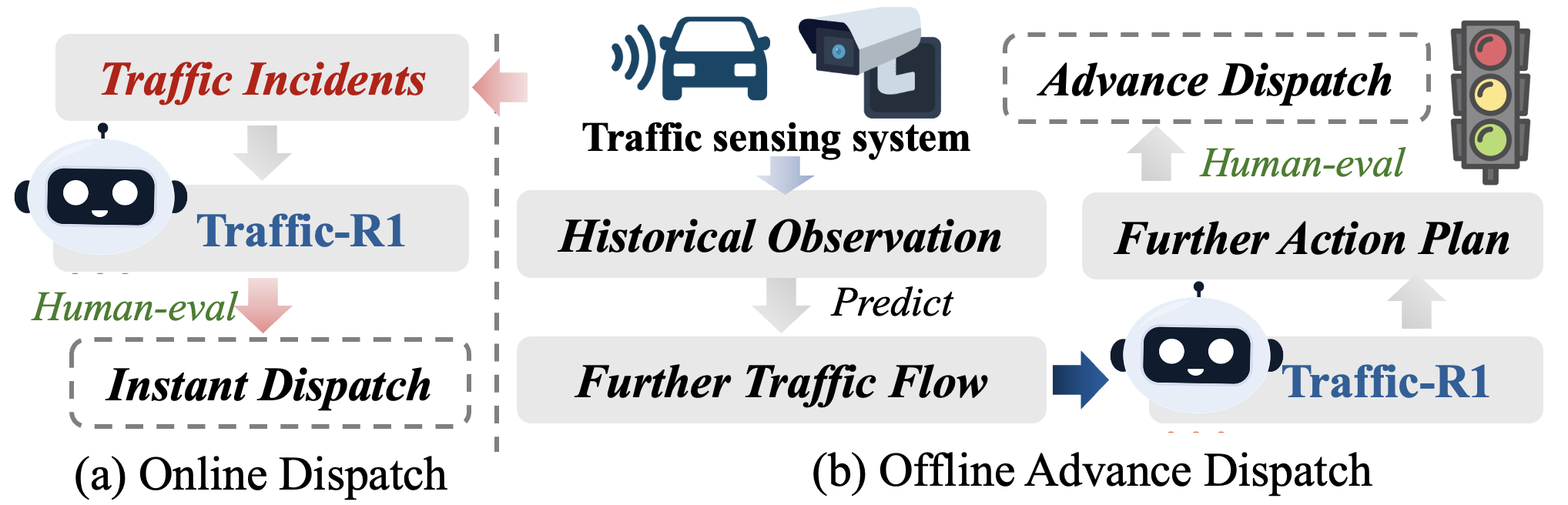}
\vspace{-2em}
\caption{Online and offline dispatch framework}
\label{fig:depoly}
\end{figure}

\subsubsection{Environment and System Adaption.}
Our model, \model, has been deployed on the official traffic management platform of a Tier-1 city in China, where it oversees 10 critical, interconnected intersections. This network serves over 55,000 vehicles daily and is situated in a central commercial district characterized by high-volume, dynamic traffic patterns. The intersections are configured as four-way junctions on arterial roads, with inter-intersection distances ranging from 600 to 800 meters. The traffic load is substantial, with peak daily throughput at a single intersection exceeding 21,000 vehicles.

To bridge the gap between simulation and real-world complexity, two key adaptations were implemented. First, to manage the increased number of lanes typical of urban arterial roads, we aggregated lanes with identical traffic movements into a single logical input, thereby reducing state-space complexity for the model. Second, the system incorporates a mandatory pedestrian-crossing phase during morning and evening peak hours. This regulatory requirement introduces additional constraints to the signal timing plan, demanding greater adaptive capabilities from our model to maintain traffic efficiency.
Real-world deployment necessitates two critical components: a multi-modal sensing system and a safety-compliant dispatch framework.

To provide the model with real-time, LLM-compatible inputs, we developed a multi-modal sensing suite. As illustrated in Figure~\ref{fig:device}, roadside cameras provide 2D visual data, which is processed by a fine-tuned Grounding DINO model for vehicle detection and classification. This is augmented by millimeter-wave radar data that provides precise vehicle-to-intersection distance measurements, creating a comprehensive traffic state representation.

Direct, fully online model control is infeasible in public infrastructure due to regulatory and safety mandates. To address this, we designed a hybrid online-offline dispatch framework (Figure~\ref{fig:depoly}). The dispatch framework is executed on a centralized server equipped with an  NVIDIA Tesla T40 24G GPU with an average inference latency of less than 2 seconds (437 output tokens in average). This centralized deployment is optimal for our predominantly offline dispatch strategy, allowing for robust data aggregation and computation. 
For routine traffic management, the system operates offline. Historical traffic data spanning 30 days is used to train a LightGBM model with spatio-temporal embeddings to forecast traffic flow for the subsequent day. Based on these forecasts, \model generates a preliminary signal timing plan. Crucially, this plan is submitted to human experts at the traffic management authority for review and final approval before implementation. The online pipeline is reserved for real-time incident response, allowing the model to make immediate adjustments under expert supervision.
This dual-pipeline approach ensures both operational efficiency and compliance with public safety standards.
Furthermore, while the current deployment leverages a center server, the model is designed for future scalability. The 3B-parameter Qwen architecture of \model can be quantized for efficient execution on low-power edge devices that lack dedicated GPUs. 

\begin{table}[h]
\centering
%\vspace{-1em}
\caption{Real-world A/B test results spanning 6 weeks.}
\vspace{-1em}
\label{tab:deploy}
\resizebox{\columnwidth}{!}{%
\begin{tabular}{cccc}
\hline
\textbf{Method} & Average Queue$\downarrow$ & Maximum Queue$\downarrow$ & Working Hours$\downarrow$ \\ \hline
Manual & 34.5 & 50.3 & 2+ \\
Traffic-R1 & 31.3 & 48.1 & 0.5+ \\
\textbf{\#Improve} & \textbf{9.3\%} & \textbf{4.4\%} & \textbf{$\sim$75\%} \\ \hline
\end{tabular}%
}
%\vspace{-1em}
\end{table}

\subsubsection{In-Situ Performance Evaluation}
We conducted a six-week in-situ evaluation to quantify the performance of \model. The experiment was structured as an A/B test, comparing our proposed system against the baseline, which consists of manual signal control by human operators. The two systems were deployed on an alternating weekly basis to mitigate temporal bias. Performance metrics, including average queue length, maximum queue length, and daily human operational hours, were recorded during weekday evening peak hours (16:00-19:30). Data from weekends and public holidays were excluded to ensure a consistent basis for comparison.

As detailed in Table~\ref{tab:deploy}, our system demonstrates marked improvements over the manual baseline. Specifically, \model reduced the average and maximum queue lengths by 9.3\% and 4.4\%, respectively. Furthermore, it achieved a substantial 75\% reduction in the human labor required for daily operations. These results empirically validate the effectiveness and efficiency of our model in a real-world, dynamic traffic environment.

% 请在导言区确保有: \usepackage{algorithm} \usepackage{algpseudocode}
\begin{algorithm*}[t]
\caption{Asynchronous TSC Network with Symmetric Half-Step Communication}
\label{alg:async_symmetric_professional}
\begin{algorithmic}[1]
\State \textbf{Input:} Road network graph $G=(\mathcal{I}, E)$, partitioned intersection sets $(\mathcal{G}_1, \mathcal{G}_2)$.
\State \textbf{Initialize:} Environment $Env$, Agent policy $\pi$, message buffer $\mathcal{M}$.

\For{each time\_step $t$ in Total\_Steps}
    \Statex \textit{// --- Half-step for agents in $\mathcal{G}_1$ ---}
    \State Create new empty buffer $\mathcal{M}_{next}$.
    \For{each intersection $i \in \mathcal{G}_1$}
        \State $O_i, M_{in, i} \gets Env.get\_observation(i), \mathcal{M}[i]$ \Comment{Read messages from $\mathcal{G}_2$ from prev. half-step}
        \State $A_i, M_{out, i} \gets \pi(O_i, M_{in, i})$
        \If{$M_{out, i} \neq \text{null}$}
            \For{each neighbor $j \in \mathcal{N}(i) \cap \mathcal{G}_2$}
                \If{$dist(i, j) \le 2\text{km}$}
                    \State $\mathcal{M}_{next}.add(j, M_{out, i})$
                \EndIf
            \EndFor
        \EndIf
    \EndFor
    \State $\mathcal{M} \gets \mathcal{M}_{next}$ \Comment{Update message buffer for the next half-step (for $\mathcal{G}_2$)}
    \State $Env.update\_state(\{A_i\}_{i \in \mathcal{G}_1})$

    \Statex \textit{// --- Half-step for agents in $\mathcal{G}_2$ ---}
    \State Create new empty buffer $\mathcal{M}_{next}$.
    \For{each intersection $j \in \mathcal{G}_2$}
        \State $O_j, M_{in, j} \gets Env.get\_observation(j), \mathcal{M}[j]$ \Comment{Read messages from $\mathcal{G}_1$ from prev. half-step}
        \State $A_j, M_{out, j} \gets \pi(O_j, M_{in, j})$
        \If{$M_{out, j} \neq \text{null}$}
            \For{each neighbor $i \in \mathcal{N}(j) \cap \mathcal{G}_1$}
                 \If{$dist(j, i) \le 2\text{km}$}
                    \State $\mathcal{M}_{next}.add(i, M_{out, j})$
                 \EndIf
            \EndFor
        \EndIf
    \EndFor
    \State $\mathcal{M} \gets \mathcal{M}_{next}$ \Comment{Update message buffer for the next half-step (for $\mathcal{G}_1$)}
    \State $Env.update\_state(\{A_j\}_{j \in \mathcal{G}_2})$
\EndFor
\end{algorithmic}
\end{algorithm*}

\end{document}